\documentclass[conference]{IEEEtran}
\IEEEoverridecommandlockouts

\usepackage{cite}
\usepackage{amsmath,amssymb,amsfonts}
\usepackage{algorithmic}
\usepackage{graphicx}
\usepackage{textcomp}
\usepackage{booktabs}
\usepackage{multirow}
\usepackage{array}
\usepackage{xcolor}
\usepackage{url}
\usepackage{hyperref}
\usepackage{tabularx}
\usepackage{siunitx}

\definecolor{dark-red}{rgb}{0.4,0.15,0.15}
\definecolor{dark-blue}{rgb}{0.15,0.15,0.8}
\definecolor{medium-blue}{rgb}{0,0,0.5}
\definecolor{draft-orange}{rgb}{0.85,0.35,0.0}
\hypersetup{
  colorlinks,
  linkcolor={dark-red},
  citecolor={dark-blue},
  urlcolor={medium-blue}
}

\def\BibTeX{{\rm B\kern-.05em{\sc i\kern-.025em b}\kern-.08em
  T\kern-.1667em\lower.7ex\hbox{E}\kern-.125emX}}

\graphicspath{{./}{res/}}
\begin{document}

% Working-title alternatives:
% 1. Seeing Beyond Daylight: Condition-conditioned RGB--Thermal--Radar Perception for ASVs
% 2. Beyond Daylight: Reliability-Aware Multimodal Perception for Small ASVs
% Avoid ``weather-agnostic'' unless rain/fog experiments support that claim.
% \title{Condition-Conditioned RGB--Thermal--Radar
% Obstacle Perception for Resource-Constrained ASVs\\
% \thanks{%
% \textsuperscript{\textdagger}Pranav Kedia and Edwin Redhead contributed equally to this work.\par
% \textsuperscript{1}Department of Computer and Information Science,
% University of Konstanz, Germany.\par
% \textsuperscript{2}Centre for the Advanced Study of Collective behavior,
% University of Konstanz, Germany.\par
% \textsuperscript{3}University College London, London, UK.\par
% {\tt\{pranav.kedia, heiko.hamann\}@uni-konstanz.de}\\
% Corresponding author: Pranav Kedia.%
% }%
% }

% \author{%
% Pranav Kedia\textsuperscript{1,2,\textdagger}, Edwin Redhead\textsuperscript{3,\textdagger}, and Heiko Hamann\textsuperscript{1,2}%
% }
\title{Context-Aware RGB--Thermal--Radar
Obstacle Perception for Resource-Constrained ASVs\\
% \thanks{The first two authors contributed equally to this work.\par
% Code, evaluation tools, additional figures, etc. are available (anonymously): \url{https://anonymous.4open.science/r/marine-multimodal-perception-B5EF/}}%
% }
% \author{%
% \IEEEauthorblockN{Anonymous Author(s)}
% \IEEEauthorblockA{Affiliations omitted for double-anonymous review}%
% }

% \title{Condition-Conditioned RGB--Thermal--Radar
% Obstacle Perception for Resource-Constrained ASVs\\
% % \thanks{The first two authors contributed equally to this work.}%
\thanks{%
\textsuperscript{\textdagger}Pranav Kedia and Edwin Redhead contributed equally to this work.\par
\textsuperscript{1}Department of Computer and Information Science,
University of Konstanz, Germany.\par
\textsuperscript{2}Centre for the Advanced Study of Collective behavior,
University of Konstanz, Germany.\par
\textsuperscript{3}University College London, London, UK.\par
{\tt\{pranav.kedia, heiko.hamann\}@uni-konstanz.de}\par
Corresponding author: Pranav Kedia.\par
% Code, evaluation tools, additional figures, etc. are available (anonymously): \url{https://anonymous.4open.science/r/marine-multimodal-perception-B5EF/}
}%
}%

\author{%
Pranav Kedia\textsuperscript{1,2,\textdagger}, Edwin Redhead\textsuperscript{3,\textdagger}, and Heiko Hamann\textsuperscript{1,2}%
}
% \author{%
% \IEEEauthorblockN{Anonymous Author(s)}
% \IEEEauthorblockA{Affiliations omitted for double-anonymous review}%
% }

\maketitle

\begin{abstract}
Reliable obstacle perception is challenging for small autonomous surface vessels (ASVs): reflections and glare confound visible-light cameras, vessel motion shifts the apparent waterline, and darkness removes the dominant sensing modality altogether. Rather than treating multimodal fusion as a static accuracy gain, we propose a context-aware architecture that combines RGB and
long-wave infrared (LWIR) imagery, frequency-modulated continuous-wave
(FMCW) mmWave radar, and inertial context. 
%We present condition-adaptive fusion of RGB imagery, long-wave infrared (LWIR) imagery, and frequency-modulated continuous-wave (FMCW) mmWave radar, with inertial context, using a bearing-sector interface for embedded navigation. 
%The representation distinguishes missing measurements from valid observations with no detections and provides a navigation controller with probability, range, and sensor-health information. 
A bearing-sector interface
provides the navigation controller with calibrated obstacle probabilities,
radar ranges, and sensor status. 
%It also distinguishes unavailable measurements from valid observations with no detections. 
It also distinguishes sensor dropout from a valid observation with no detected obstacle. 
%The complete pipeline, from new multimodal lake recordings to an onboard fusion model, is demonstrated in closed-loop navigation trials under uncontrolled lake conditions, including night runs in which the camera returns no usable evidence and the vessel continues to avoid obstacles on LWIR and radar alone. 
Using new multimodal lake recordings, we deployed the pipeline in a 24-leg closed-loop pilot under uncontrolled
lake conditions (16~day, 4~dusk, and 4~night; \SI{351}{m}), including
night legs in which RGB was unusable and the vessel drove on LWIR and radar evidence alone. 
On eight held-out low-light recordings, the calibrated fusion model achieved average precision (AP) 0.869 and expected calibration error 0.108, compared with AP 0.673 for rule-based fusion, 0.620 for gated fusion, and 0.903 for RGB alone. Our results expose a central tradeoff: 
RGB ranks best when usable, whereas fusion preserves a navigation input when RGB fails.
%In a closed-loop pilot, the vessel drove 24 legs (16 day, 4 dusk, 4 night; \SI{351}{m}) on that stream. 
This work is a step toward perception that remains usable and degrades predictably across multi-day ASV missions.
\end{abstract}


\section{Introduction}
Small ASVs are attractive for environmental monitoring and long-duration observation (e.g., for several days and nights) because they can operate in water with low logistical cost. On inland water bodies, however, their safety and autonomy depend on detecting obstacles that range from boats and fixed structures to swimmers, animals, buoys, and partly submerged debris. This is a difficult visual domain. Sun glitter and wakes can resemble physical obstacles, and the horizon moves with roll and pitch~\cite{zust2023lars}. At night, the visible-light channel that many perception systems on small ASVs rely on may cease to be useful. The challenge is to interpret changing sensor evidence while preserving the information needed for robot decisions. Long-duration field autonomy requires sustaining task performance without exhausting energy, and on an ASV that budget must cover both perception (sensors and computation) and propulsion~\cite{egerstedt2021robot}. 

Autonomous wind-powered surface vessels (ASWVs)~\cite{kedia2024onyx} can reduce electrical propulsion demand, while sensing, computation, communication, and control continue to draw battery power. Perception must therefore deliver timely, decision-relevant obstacle information across day and night, from sensors whose reliability and availability vary, and within the energy and compute budget of a small vessel. No single compact sensor resolves all of these challenges. RGB imagery provides texture and semantic class information in daylight, but is sensitive to illumination. Long-wave infrared (LWIR) imaging retains contrast for warm objects in darkness, but has lower resolution and weak contrast for thermally uniform structures. FMCW mmWave radar provides range and radial velocity independently of visible illumination, but sparse returns and target-dependent reflectivity complicate interpretation. Inertial measurements provide no obstacle evidence, but they describe the viewing geometry and platform motion under which the other sensors observed the scene.  A measurement's value therefore depends on its acquisition conditions: illumination, roll and pitch, and sensor availability (dropouts, field-of-view gaps, unresolvable sectors). We study whether explicitly representing these conditions, rather than uniformly fusing sensor outputs, yields better-calibrated and more robust obstacle estimates on an ASV.

Existing maritime work has made substantial progress in camera-based water-obstacle segmentation~\cite{tersek2023ewasr} and in radar--camera datasets and fusion~\cite{yao2024waterscenes}. Researchers have also fused electro-optical and infrared cameras for maritime tracking~\cite{park2024multivessel}. These systems show the value of multiple modalities, but a resource-constrained ASV also needs low-energy, bandwidth-efficient, navigation-specific output, explicit sensor-health handling, and evidence that the full perception-to-control chain survives severe condition changes. This paper connects session-disjoint perception evaluation to embedded runtime measurements and a closed-loop lake pilot, distinguishing offline model performance from the evidence available during robot decisions. To our knowledge, this is the first demonstration of in-field closed-loop obstacle avoidance in the marine robotics domain using joint RGB, LWIR thermal, and FMCW mmWave radar fusion with inertial context.

\begin{figure}[t]
  \centering
  \includegraphics[width=\columnwidth]{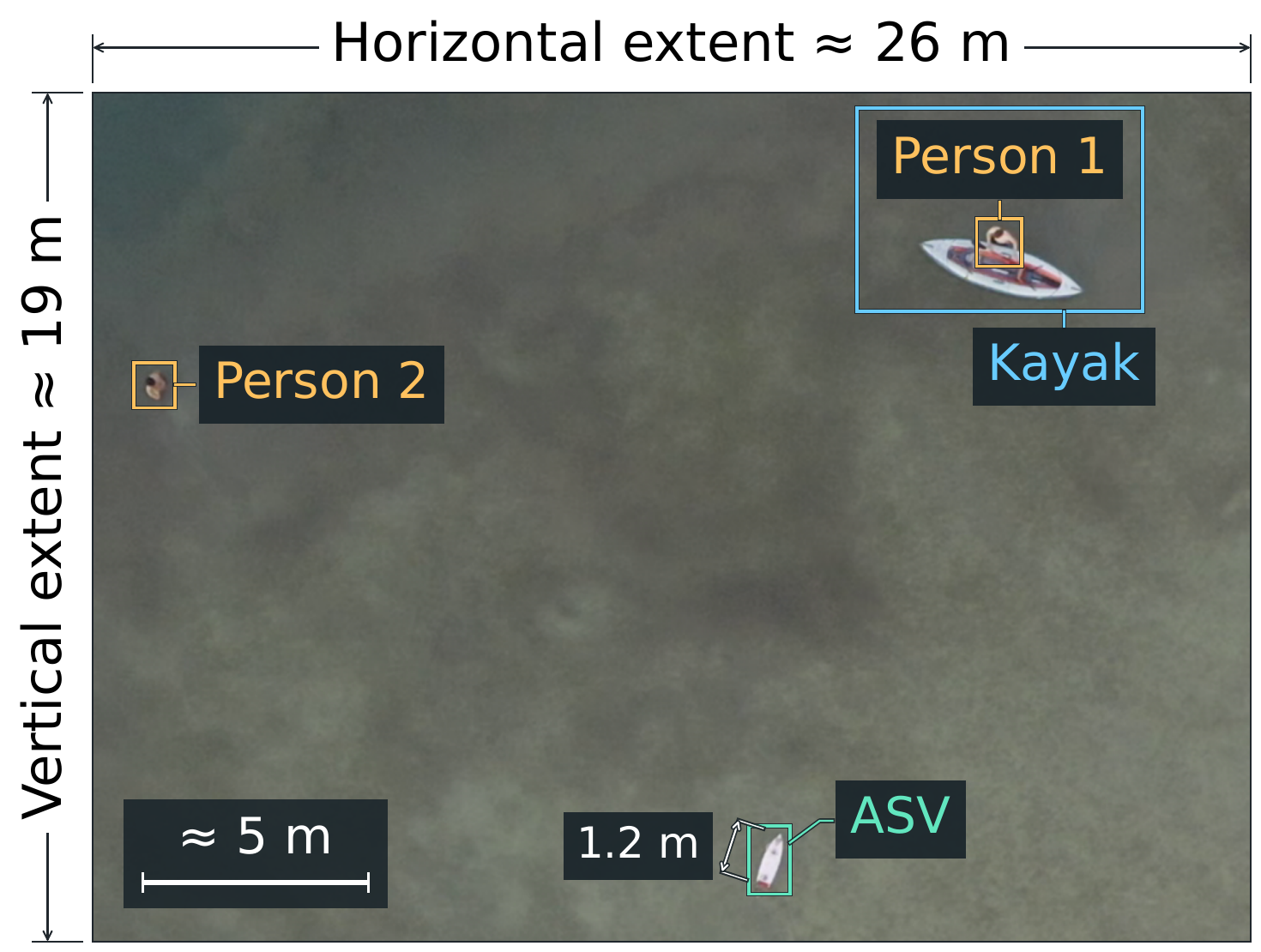}
  \caption{Aerial view of the field trial showing the ASV, kayak, and two people.}
  \label{fig:field}
\end{figure}

Our key contributions are:
(1)~A~context-conditioned RGB--LWIR--FMCW mmWave radar perception method with inertial context and a compact sector interface to navigation.
(2)~A~session-disjoint evaluation, isolating the effects of geometric compensation, contextual inference, probability calibration, and sensor degradation, and showing that at true night every learned or pseudo-labeled path must be gated on RGB luminance.
(3)~An embedded lake pilot (Fig.~\ref{fig:field}) relating the available obstacle evidence, processing configuration, and control policy to 24 driven legs across daylight, dusk, and night.

Extending small-ASV autonomy beyond daylight requires obstacle estimates that remain trustworthy as sensors degrade and drop out, and we show that this can be achieved on a single low-power embedded board.

%The closed-loop navigation evaluation uses a medium-speed, motor-assisted pilot on the AWSV platform. The controller receives the sector stream over the link described in Sec.~\ref{sec:runtime}
%and, in the final sessions, a deliberately simple policy holds station,
%chooses the least-obstructed heading from the sector probabilities, places a
%checkpoint on it and drives there under motor at medium speed
%(10 trials, Sec.~\ref{sec:closedloop}). The interface is designed for
%a future COLREG-aware planner.

\section{Related Work}
Classical maritime detection estimates the water edge and separates obstacles from navigable water~\cite{zust2023lars}. Deep segmentation improved robustness to reflections and wakes (WaSR) with inertial cues~\cite {bovcon2022wasr}, \textit{eWaSR} for embedded compute~\cite{tersek2023ewasr}, and \textit{LaRS} extends evaluation to diverse lake, river, and sea scenes~\cite{zust2023lars}. 
%We use an embedded segmentation model distilled from these models to provide visual evidence, while relying on complementary sensors when segmentation becomes unreliable in darkness.
We use an embedded segmentation model distilled from these models to provide visual cues, relying on complementary sensors when segmentation degrades in darkness. 

Multimodal sensing on ASVs is well motivated but rarely carried through to inference and control. \textit{WaterScenes} provides a benchmark demonstrating that radar and camera provide complementary information for detection, free-space segmentation, and waterline estimation on ASVs~\cite{yao2024waterscenes}, and electro-optical/infrared track fusion has been applied to maritime targets~\cite{park2024multivessel}. RGB, LWIR, mmWave radar, and inertial recordings exist for indoor and outdoor ground robots~\cite{li2022odombeyondvision,jiang2025go}, and HydraFusion studies context-dependent sensor selection for road vehicles~\cite{malawade2022hydrafusion}. These studies leave open how joint RGB, LWIR, and FMCW mmWave radar evidence supports marine robot navigation when illumination, viewing geometry, and onboard resources change. Our study examines context-adaptive obstacle estimates and their use for control under the resource and viewing-geometry constraints of a small ASV.

Creating ground truth for maritime scenes is labor-intensive: classes are diverse, instances are rare, and existing detectors recognize only a limited set of classes. Open-vocabulary detectors such as \textit{Grounding DINO}~\cite{liu2024groundingdino} and concept-prompted segmentation in \textit{SAM~3}~\cite{carion2025sam3} make it practical to generate candidate annotations from text prompts, reducing the need for maritime class-specific detectors. 
%We use these annotations as \emph{teachers} whose proposals a human audits, and we quantify their recall and precision against that audit rather than treating them as truth.
We use these annotations as \emph{teachers} whose proposals we audit, and quantify recall and precision against that audit rather than treating them as truth. 

Perception benchmarks alone do not reveal whether errors occur at safety-critical ranges or whether an embedded system reacts quickly enough for control. Radar has been used directly in ASV obstacle-avoidance systems~\cite{shi2026millimeterwave}, but radar-only control sacrifices the semantic and navigable water information available from cameras. We therefore evaluate both obstacle perception and its effectiveness in closed-loop behavior.

\section{System and Operational Overview}
Our perception module combines RGB, LWIR thermal, and FMCW mmWave radar measurements with inertial context on an embedded computer. Its output provides the navigation controller with bearing-resolved representation of detected obstacles. Next, we describe the method's operating constraints, sensor configuration, and timing/geometric references.

\subsection{Energy and Response-Time Constraints}
\label{sec:energy_design}
The perception subsystem consumes energy through sensing, computation, interfaces, and power conversion. Expressing all power flows at the battery terminals, the mission-level energy budget is
\begin{equation}
 \dot E=P_{\mathrm{harvest}}-
 (P_{\mathrm{perception}}+P_{\mathrm{aux}}+P_{\mathrm{locomotion}})\;,
 \label{eq:energy_budget}
\end{equation}
where $P_{\mathrm{aux}}$ includes navigation and communication, and $P_{\mathrm{locomotion}}$ includes actuation and any motor propulsion. The closed-loop navigation evaluation uses a medium-speed, motor-assisted pilot on the ASWV platform. Energy availability bounds mission duration, and target range and closing speed bound the allowable perception-to-action latency: for a target closing at $v_{\mathrm{close}}$ we require $t_{\mathrm{age}}+t_{\mathrm{control}}+t_{\mathrm{maneuver}}+t_{\mathrm{margin}} < R_{\mathrm{validated}}/v_{\mathrm{close}}$, where $t_{\mathrm{age}}$ is the age of the evidence when the controller receives it, so the validated sensing range and the end-to-end latency jointly bound the closing speeds a timely maneuver can handle.
Table~\ref{tab:power_design} compares illustrative sensing and compute power budgets alongside illumination and weather limitations.
These component-level estimates motivate the sensor and compute choices.

\begin{table*}[t]
\caption{Illustrative sensing-plus-host-compute budgets (in W).
%\textbf{R}: RGB camera; \textbf{T}: LWIR camera; \textbf{M}: compact mmWave radar; \textbf{L}: compact 3-D LiDAR. 
The first five rows use the same Raspberry Pi 4 compute allowance.}
\label{tab:power_design}
\centering
\footnotesize
\setlength{\tabcolsep}{4pt}
\begin{tabularx}{\textwidth}{l l >{\raggedright\arraybackslash}X r r r}
\toprule
Sensors & Unlit-night evidence & Rain/fog consideration & Sensing (W) & Compute (W) & Sum (W) \\
\midrule
\textbf{R}(GB Camera) & Ambient light required & Optical contrast/visibility limited & 1.0 to 1.5 & 3 to 7 & 4.0 to 8.5 \\
\textbf{T}(hermal LWIR) & Thermal contrast & Contrast, optics, and fog dependent & 0.3 to 1.0 & 3 to 7 & 3.3 to 8.0 \\
\textbf{M}(mWave radar) & Range and Doppler & Useful non-optical channel; clutter/weak returns & 2 to 4 & 3 to 7 & 5 to 11 \\
\textbf{R+T+M} & Thermal and radar & Candidate for wider operating coverage & 3.3 to 6.5 & 3 to 7 & 6.3 to 13.5 \\
\textbf{R+L}(iDAR 3D) & LiDAR geometry & Optical attenuation/backscatter; active laser & 7.5 to 8.0 & 3 to 7 & 10.5 to 15.0 \\
\textbf{R+T+M}, accelerated & Thermal and radar & Same sensing; different compute budget & 3.3 to 6.5 & 9 to 30 & 12.3 to 36.5 \\
\bottomrule
\end{tabularx}
\par\smallskip
\end{table*}

\subsection{Sensor Suite and Embedded Processing}
\begin{figure}[t]
    \centering
    \includegraphics[width=\columnwidth]{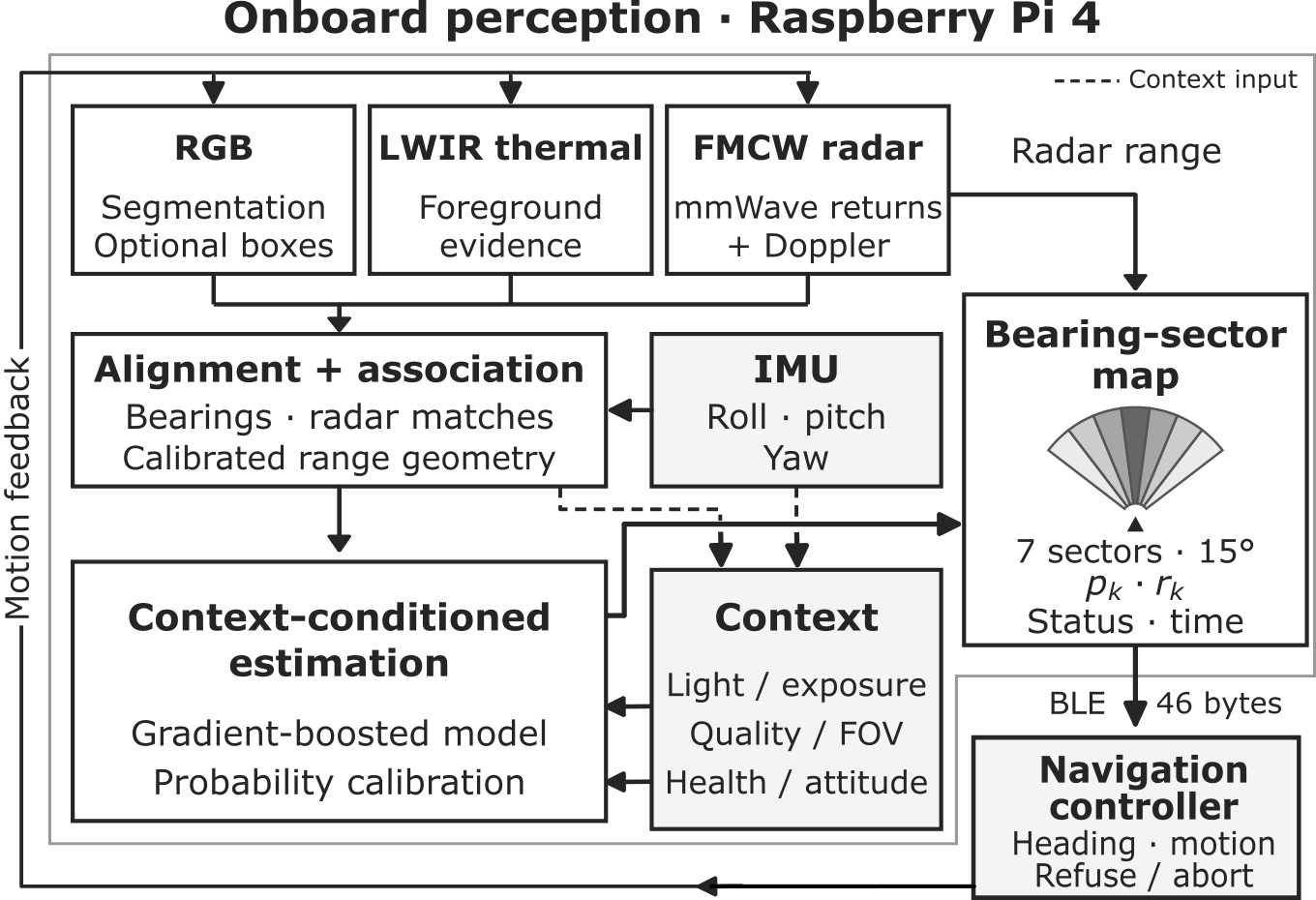}
    \caption{Overview of our multimodal perception module. Synchronized RGB, thermal, radar, and inertial measurements produce modality-specific evidence. Context-conditioned fusion converts this evidence into a compact bearing-sector representation for the navigation controller.}
    \label{fig:system_overview}
\end{figure}

Figure~\ref{fig:system_overview} summarizes our hardware and processing pipeline. The water-resistant enclosure houses the sensors and a Raspberry Pi~4, with components listed in Table~S3. The RGB fisheye camera captures detailed scene information in daylight, while LWIR thermal imaging and FMCW mmWave radar provide complementary observations when visible-light imaging is impaired. The IMU supports geometric compensation for vessel motion. A~battery-backed real-time clock provides an offline time reference, and the autopilot supplies GPS position over a serial link for computing sun elevation.
We can enable or disable each sensor stream independently. We distinguish valid observations containing no detections from unavailable measurements. For example, we retain an empty radar scan as a valid observation but mark a disabled or failed stream as unavailable. For each stream, we also record status metadata, including measurement age, rejected-packet counts, and violations of configured point-count limits. We provide the measurements and their status metadata to the fusion stage so that it can account for sensor availability, and we store this information to reproduce dropout experiments during offline replay.
We read radar packets continuously over UART, validate each packet's length, and process the radar stream independently of the camera frame rate. Before each field campaign, we verify the mounting orientations and sensor alignment. We remove returns from the vessel's own structure using the self-return filter described in Sec.~\ref{sec:radar}.

\begin{figure}[t]
  \centering
  \includegraphics[width=0.47\columnwidth]{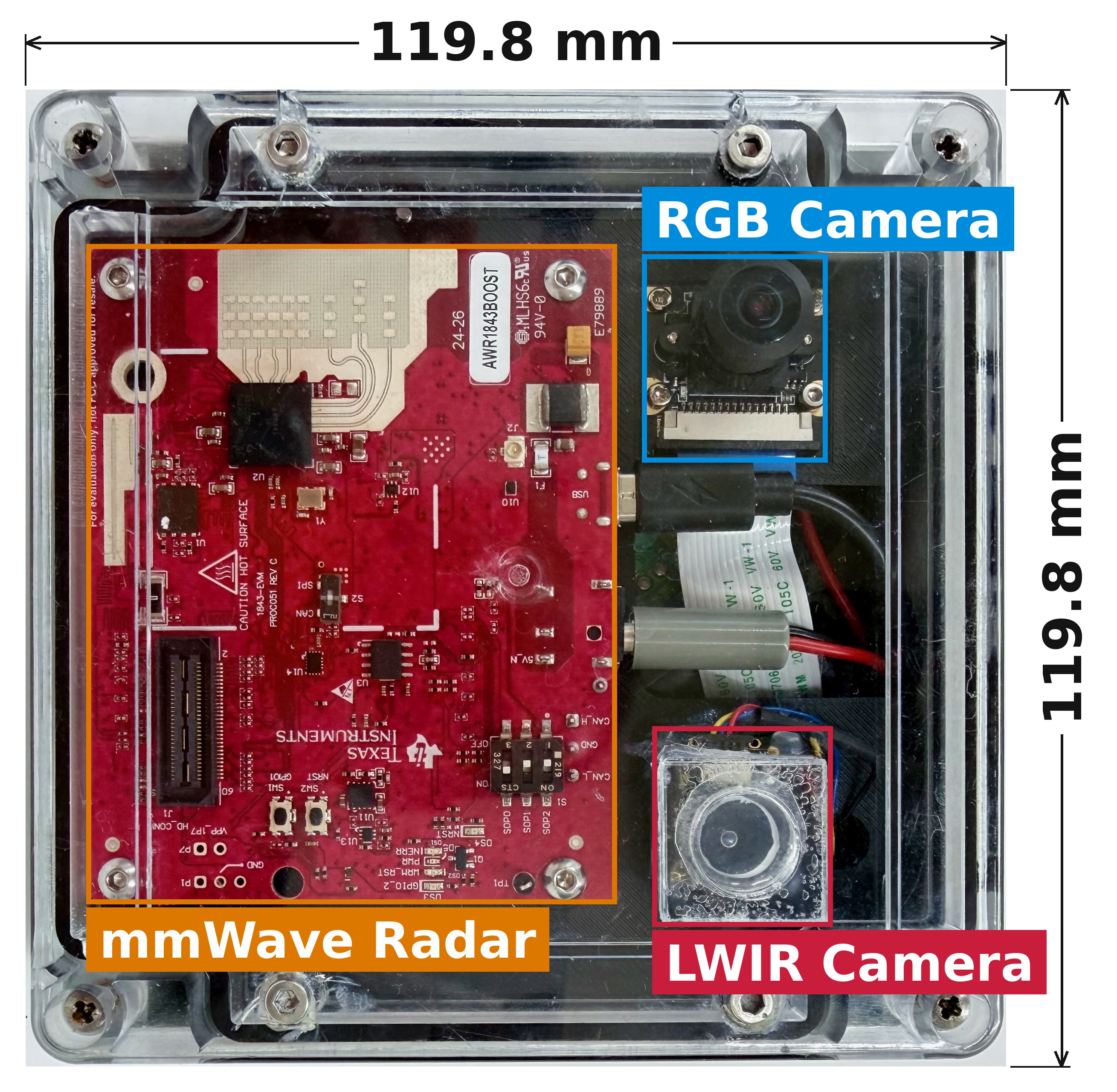}
  \includegraphics[width=0.47\columnwidth]{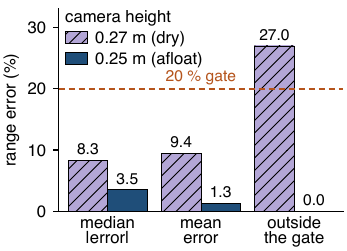}
  \caption{Left: the perception box. Right: monocular range error on an isolated wading target at 6.9--\SI{8.9}{m} before and after the afloat calibration (camera height 0.27$\rightarrow$\SI{0.25}{m}, IMU attitude, water-line): median absolute error 8.3\% $\rightarrow$ 3.5\%, 73\% $\rightarrow$ 100\% of detections within the $\pm$20\% gate ($n=11$).}
  \label{fig:boat}
\end{figure}

\subsection{Temporal Alignment and Sensor Geometry}
\label{sec:timing_calibration}
We timestamp all sensor streams using the onboard computer's system clock, which an external RTC maintains across power cycles. We divide recordings into \SI{300}{s} chunks and target a capture rate of \SI{3}{Hz}. At the start of each capture-loop iteration, we write a system-clock timestamp at \SI{0.1}{s} resolution to a separate metadata file associated with the camera frame, which we call the frame sidecar. We timestamp radar and IMU samples when the onboard computer receives or decodes them, and each capture iteration stores the latest available sensor data. During online inference, we associate each camera snapshot with its sidecar using the recording-chunk identifier and the frame index within that chunk. This procedure permits a snapshot to be one frame newer than its sidecar record, corresponding to a nominal offset of \SI{333}{ms} at \SI{3}{Hz}. When capture outpaces inference, we skip records superseded by newer data. We select the latest complete radar timestamp group and the nearest logged IMU sample within \SI{1}{s}. In the full perception configuration, we reuse asynchronous segmentation results for up to \SI{2}{s}. The typed configuration keeps masks and detections valid for up to \SI{4}{s}.

We undistort the RGB images using the calibrated fisheye model and compute bearings from the resulting pinhole geometry. For the thermal camera, we convert image position to bearing using a scale of \SI{2.82}{px} per degree, corresponding to a horizontal field of view of approximately \(56.7^{\circ}\). An independent field calibration yielded \(58.2^{\circ}\) (95\% CI: $[52.8^{\circ},68.6^{\circ}]$), supporting our use of this mapping. We measured the sensor extrinsics photogrammetrically, with approximately \(\pm\SI{1}{mm}\) in-plane uncertainty and \(\pm\SI{2}{mm}\) depth uncertainty. From a walk test (ASV on land with trolley), we estimated a radar yaw offset of \(-6.0^{\circ}\), which we correct before associating measurements and assigning them to sectors. We define the local coordinate frame with~$x$ pointing right, $y$~forward, and $z$~upward. Radar bearing is \(\theta=\operatorname{atan2}(x,y)\), with positive angles to the right. We express all sector bearings in this local frame. For monocular ranging, we estimate the water-plane orientation from the segmented water edge when available. Otherwise, we use, in order, the IMU attitude, the detected horizon, or an assumed level orientation. We measured the RGB-camera height afloat as \SI{0.250}{m} and derived a thermal-camera height of \SI{0.188}{m}. Whenever we remount the sensors, we remeasure these heights and re-estimate the radar self-return region.

\subsection{Bearing-Sector Map and Navigation Interface}
\label{sec:interface}
We partition the forward-facing angular region into a bearing-indexed map of $K=7$ contiguous $15^{\circ}$ sectors, centered at $-45^{\circ}, -30^{\circ}, \ldots, +45^{\circ}$ and spanning $\pm52.5^{\circ}$. For each sector~$k$, we estimate a calibrated obstacle probability~$p_k$ and report the minimum radar-derived forward distance~$r_k$. We encode an unknown distance explicitly as null. This fixed-size sector representation provides a common interface between multimodal perception and navigation. Depending on the processing configuration, we also generate diagnostic outputs such as cross-modal confirmation, closing-speed and time-to-collision estimates, vessel attitude, navigable-water profiles, and tracked-target velocities and closest-point-of-approach predictions.

We transmit a 46-byte packet to the navigation computer over Bluetooth Low Energy. The packet contains sector scores, calibrated probabilities, distances, and frame-level status flags, together with a sequence number, a~timestamp at \SI{0.1}{s} resolution, processing time, and source latency when available. The status flags indicate whether probability estimation succeeded and whether the segmentation and typed detections remain fresh. We preserve unknown distances through an explicit missing-value encoding. We do not transmit target tracks or velocity estimates through this interface. We also omit predictions of when and how closely a tracked target will pass the vessel. The controller bases its decisions on the transmitted sector estimates. The sensing and inference configuration determines the update rate and age of these estimates (Sec.~\ref{sec:runtime}).

\section{Multimodal Perception and Sector Fusion}
\subsection{RGB Segmentation and Ranging}
We first undistort each RGB frame and use an embedded segmentation network to classify pixels as obstacle, water, or sky. We trained an LR-ASPP/MobileNetV3 student network through knowledge distillation~\cite{hinton2015distilling} from \textit{eWaSR}~\cite{tersek2023ewasr}, using \textit{LaRS}~\cite{zust2023lars} and 10{,}731 frames from our own recordings labeled by the teacher. The resulting \SI{3}{MB} MobileNetV3 model~\cite{howard2019mobilenetv3} agrees with the teacher on 99.5\% of pixels across the distillation corpus; this measures teacher reproduction rather than accuracy against independently audited labels. On the onboard computer, the \textit{int8} student processes $512\times384$ images at \SI{4.5}{fps}, compared with \SI{1.3}{fps} for the teacher.

For each segmented obstacle, we select the lowest pixel next to predicted water as the waterline contact point. Combined with camera height and attitude, this estimates monocular range (Fig.~\ref{fig:boat}). We flag detections in predicted water without a contact point as possible reflections or glints. The fusion model uses mean luminance, exposure time, gain, and predicted water fraction as context features rather than general thresholds. At \SI{3}{fps}, automatic exposure can reach \SI{300}{ms} at dusk, allowing near-darkness to appear bright. We reject frames with mean luminance below 25. During each night trial segment (mean luminance 0.1–1.7), the student labels 99.6–100\% of the frame as obstacle, the open-vocabulary teacher emits boxes, and an offline replay using those masks disagreed with vessel records until we applied the same gate. We mark such masks unavailable to fusion, and use the gate for training and scoring.

\subsection{LWIR Obstacle Detection and Water Segmentation}
We correct each LWIR frame for its fixed mounting orientation and map image columns to bearings using the field-validated linear calibration. We extract foreground candidates using a running-mean background and local contrast. On audited night sequences, this detector found 51 of 53 persons and 383 of 409 boats, with 0.89 recall and 0.52 precision. MOG2, although best on daytime proxy data, achieved only 0.53 recall against the same annotations because the night scenes were largely static. On the trial evening's 247 audited true-night frames (RGB black, truth drawn on the thermal image), the same detector recalls 100 of 122 persons, 40 of 46 boats, and 22 of 23 structures (0.85) at strict precision 0.17, with 3.6 false alarms per frame on water texture: thermal sees what is there at night, and absorbing its false alarms is the fusion's job. We also run a thermal segmentation student trained jointly on grayscale RGB priors and warped thermal labels. It processes images at \SI{25}{fps} on the onboard computer and achieves 0.915 pixel accuracy on a held-out under-pier scene. We use its predicted water boundary and navigable-water region. Because LWIR provides neither reliable semantic contrast across all objects nor depth, we allow thermal detections to increase obstacle probability but do not derive range from them.

\subsection{Radar Filtering and Relative-Motion Tracking}
\label{sec:radar}
For each radar point $(x,y,z,v)$ with SNR, where $v$ is radial Doppler velocity, we compute $r=\sqrt{x^2+y^2}$ and $\theta=\operatorname{atan2}(x,y)$. We filter points by range and height, and suppress returns within a measured self-clutter zone. A~Doppler threshold preserves moving returns. We then compute the nearest range and closing speed in each sector. After every remount, we remeasure the self-clutter zone from an occupancy map of recorded returns. For the current mount, we raised the Doppler threshold from \SI{0.15}{m/s} to \SI{0.30}{m/s} to reject two-bin noise and extended the zone to port, filtering one-fifth of near-return sectors without changing retained sectors. On the audited holdout, they reduced AP by 0.017, within the variation across recordings. We cluster radar points across frames and estimate bearing rate over a \SI{2}{s} window. For each track, we combine tangential velocity $r\dot{\theta}$ with radial Doppler velocity to estimate 2-D relative velocity in the boat frame, classify the motion as closing, crossing, or diverging, and predict the closest point of approach. We compensate for ego rotation using IMU yaw but retain ego translation because collision geometry depends on relative motion.\par

\subsection{IMU-Stabilized Association}
\label{sec:assoc}
We project radar returns into each undistorted camera image using the measured extrinsics. We accept a match only if the return lies within the detection's angular support and its sector receives a vote from the matched returns. We reject returns on water or unmasked pixels at less than half the camera-estimated range as nearer foreground objects admitted by the pixel gate. Radar confirmation thus requires directional agreement, reducing confirmed bins without a same-sector radar vote from 26\% to 0\%. If the camera-estimated target lies within radar coverage but its box contains no return, we withhold confirmation.

\subsection{Context-Conditioned Fusion}
Let $\mathbf{e}_k=[e^{\mathrm{rgb}}_k,e^{\mathrm{th}}_k,e^{\mathrm{rad}}_k]$ be the modality evidence for sector~$k$ (blob and segmentation occupancy, thermal occupancy, radar hits, nearest range, SNR statistics, confirmation) and $\mathbf{c}_k$ the context vector: sun elevation from RTC time and position, luminance and exposure, thermal-frame contrast, radar noise floor, attitude, range band, field-of-view membership and health flags. We compare three estimators of $p_k$ on identical inputs: (i)~the rule-based vote $n_{\mathrm{hits}}/3$ that the system started with; (ii)~an interpretable non-negative gate,
\begin{align}
  w_{k,s} &= m_{k,s}\,\operatorname{softplus}(u_s+\mathbf{v}_s^{\top}\mathbf{c}_k)\;, \\
  z_k &= b(\mathbf{c}_k)+\sum_s w_{k,s}\,e_{k,s}\;, \qquad \hat p_k=\sigma(z_k)\;,
\end{align}
where $m_{k,s}\in\{0,1\}$ is the availability mask, so evidence can never become anti-evidence; and (iii)~a~gradient-boosted classifier~\cite{ke2017lightgbm,pedregosa2011sklearn} over $[\mathbf{e}_k,\mathbf{c}_k]$ that can represent cross-modal interactions that the additive gate cannot. We calibrate all three estimators using isotonic regression~\cite{zadrozny2002isotonic} and package the calibration map with the deployed artifact.
%Calibration is necessary because an uncalibrated, class-balanced model assigned $p\in[0.4,0.7]$ to open water without obstacle indications. 
We calibrate the probabilities to correct distortions from class-balanced training. We fit the map to out-of-fold predictions using whole recordings as folds.
%Our first deployed bundle instead fitted it to training predictions, worsening ECE from 0.23 to 0.32; out-of-fold calibration achieved 0.11 (Table~\ref{tab:ablation}), so we replaced the bundle. 
Out-of-fold calibration reduced ECE from 0.23 to 0.11 (Table~\ref{tab:ablation}). Each training row represents one frame--sector pair. We derive targets from audited boxes projected to bearing, with a half-sector angular tolerance and $\pm7$-frame dilation to reduce label flicker. We split data by recording session, never by frame.

\subsection{Perception-to-Control Interface}
The navigation controller receives sector probabilities, radar-derived distances, and frame-level status (Sec.~\ref{sec:interface}). It uses an obstacle-evidence score  $q_k$ for heading selection, with larger values indicating stronger evidence of an obstacle in sector~$k$. Initially, $q_k=p_k$. In later trials, we used $q_k=\max(p_k,v_k)$, where $v_k=n_{\mathrm{hits},k}/3$ is the fraction of sensing modalities reporting an obstacle in that sector. This rule retains sensor agreement when the learned probability is low. The navigation policy and changes to its abort rule are reported in Sec.~\ref{sec:closedloop}.

%%%

\section{Data, Labels and Experimental Design}
\subsection{Field Site, Sessions and Corpus}
Recordings were made on a lake (Fig.~\ref{fig:field}) from our ASV, a pontoon, and a moored vessel, between June and September 2026: six daylight, two twilight, and one night session (\SI{13.0}{h}, 121{,}313 synchronized frames in the curated corpus), with buoys, moored and moving boats, wading and swimming people, water birds, and shoreline structures at ranges from \SI{1}{m} to the shore. Illumination is measured: exposure time, gain, and luminance are logged per frame, and sun elevation is computed from RTC time and GPS position. Rain footage exists only from an early prototype and is not used quantitatively.

\subsection{Teacher-Audited Labels}
Two open-vocabulary teachers proposed boxes on undistorted frames, which a human audited using a browser tool. Set~A (698 frames, 9,046 boxes) includes 350 low-light, 150 day, and 198 negative frames (left empty by both teachers); Set~B adds 300 validation frames; tranche~C (400 frames) is reserved for Sec.~\ref{sec:results}. Sets index audit batches; the train–test split is by session. Against the audit (IoU 0.5, class-aware), SAM~3 prompts \cite{carion2025sam3} achieve 0.96 recall in day and low light at 0.85–0.88 precision, versus 0.29 and 0.20 recall at 0.97–0.99 precision for Grounding DINO \cite{liu2024groundingdino}. SAM~3 seeds all labels; DINO is a second opinion. Audit-from-suggestion favors the teacher proposing more boxes, so recall is an upper bound. The audit recovered 444 small, distant objects missed by both teachers on the 198 empty frames. Suppressor prompts for teacher failure modes (night “person” switched on the outboard motors, $P=0.48$; water as “duck”, $P=0.18$) failed an A/B test: any prompt reduced joint scoring (recall $0.94\rightarrow0.71$). A geometric reflection filter on the water mask removed 264 false positives at a cost of 13 true boxes, and a YOLOv8s detector fine-tuned on audits rejects 96\% of motor-as-person proposals while keeping 82\% of true persons.

\subsection{Baselines, Ablations and Metrics}
Offline replay ensures every comparison uses identical measurements. We compare RGB-, thermal- and radar-only evidence, the rule-based vote, the non-negative gate and the gradient-boosted model. Ablations remove the IMU attitude reference, radar association, Doppler/target motion, and calibration. Degradation tests replay recorded sensor dropouts, a misted thermal window (day~1: 73\% of thermal frames below the contrast floor), timestamp offsets, and the radar's static-clutter filter. Perception is evaluated at the sector interface: precision, recall, F1, and average precision (AP) stratified by condition, range band, and class; false-positive sectors per minute on obstacle-free open water, Brier score and expected calibration error (ECE) \cite{guo2017calibration}, range error where radar truth exists, and the longest false-negative gap per encounter. We measure runtime metrics on the board, with the CPU clock and SoC temperature sampled around each iteration.

\section{Results}
\label{sec:results}

\subsection{Calibration and Range}
After the afloat calibration (camera height re-measured under load, IMU attitude, water-line contact point), monocular range on an isolated wading target at 6.9 to \SI{8.9}{m} has a median absolute error of \SI{3.5}{\percent} (mean $+1.3\%$, 100\% within $\pm20\%$, $n=11$), against 8.3\%/73\% at the out-of-water height (Fig.~\ref{fig:boat}). The height correction accounted for most of the gain. Over whole dockside clips, the error stays large because the camera sees the shore beyond the radar's \SI{9}{m} range. We introduced a range hierarchy (Sec.~\ref{sec:assoc}). 

% \result{Range MAE vs.\ GPS-tracked targets from the final sessions: XX m at XX--XX m.}

\label{sec:calibration_results}
We checked the calibration in the field. The in-field estimate of the LWIR horizontal field of view was $58.2^{\circ}$ (95\% CI [$52.8^{\circ}$, $68.6^{\circ}$]), consistent with the nominal value in Table~S3, whereas a checkerboard calibration had yielded $103^{\circ}$. Projecting radar returns from a reference target into the camera image yielded \SI{10.8}{px}, compared with \SI{10.6}{px} predicted (the image offset of a dead-ahead radar return from the principal point, a reprojection residual of \SI{0.2}{px}). The walk-test fit gave a radar mounting yaw of $-6.0^{\circ}\pm0.8^{\circ}$ (half the range of four independent estimates) and a bearing scale of 0.999. Over 14 rocking cycles of $\pm13^{\circ}$, IMU pitch and roll correlated with the corresponding water-edge measurements ($r=+1.00$ and $r=+0.94$, respectively).

\subsection{Target Motion}
On a crossing encounter (a wading subject transiting $-45^{\circ}$ to $+10^{\circ}$ at 6 to \SI{8.7}{m}, Doppler near zero mid-crossing) the tracker classifies \emph{crossing} with precision 1.00 and recall 0.81; on a deliberately rocked static sequence it produces zero false crossings, with a tangential-velocity floor of \SI{0.145}{m/s} (95th percentile) against a crossing signal of \SI{0.54}{m/s}. 

% \result{Boat crossing and sustained-turn encounters from the final sessions: XX/XX correct states, ego-yaw correction on/off.}

\subsection{Perception Across Illumination}
Table~\ref{tab:perception-results} reports performance by illumination on the held-out September session. At night, the rule-based vote achieves AP~0.58, compared with at most~0.52 for the tested learned models. Thermal-only and radar-only models, however, recall 0.90 and 1.00 of the audited persons at $p\geq0.5$, against 0.17 for the vote: the vote ranks sectors better overall, but with RGB rejected it needs thermal and radar to agree, and a person is usually seen by only one of them.

\begin{table}[t]
\caption{Held-out September 8 results: 532 day, 917 twilight, and 1{,}561 night sectors. Models were trained on six earlier sessions; night labels include 100 frames with thermal-box annotations. Lower Brier and ECE are better.}
\label{tab:perception-results}
\centering
\footnotesize
\setlength{\tabcolsep}{1.5pt}
\begin{tabular}{@{}l*{9}{r}@{}}
\toprule
& \multicolumn{3}{c}{Day}
& \multicolumn{3}{c}{Twilight}
& \multicolumn{3}{c}{Night} \\
\cmidrule(lr){2-4}
\cmidrule(lr){5-7}
\cmidrule(lr){8-10}
Method & AP & Brier & ECE
       & AP & Brier & ECE
       & AP & Brier & ECE \\
\midrule
RGB only
& 0.907 & 0.215 & 0.278
& 0.766 & 0.259 & 0.209
& 0.453 & 0.289 & 0.208 \\
Thermal only
& 0.896 & 0.305 & 0.399
& 0.811 & 0.288 & 0.295
& 0.453 & 0.273 & 0.127 \\
Radar only
& 0.915 & 0.160 & 0.213
& \textbf{0.822} & 0.200 & 0.073
& 0.450 & 0.322 & 0.238 \\
Rule-based vote
& \textbf{0.919} & 0.339 & 0.441
& 0.817 & 0.419 & 0.439
& \textbf{0.582} & 0.290 & 0.220 \\
Non-negative gate
& 0.853 & 0.281 & 0.358
& 0.774 & 0.215 & 0.114
& 0.425 & 0.377 & 0.357 \\
\textbf{GBT with context}
& 0.898 & 0.395 & 0.513
& 0.798 & 0.287 & 0.261
& 0.516 & 0.298 & 0.239 \\
\bottomrule
\end{tabular}
\end{table}

\subsection{Modality Handover and Degradation}
\begin{figure}[t]
  \centering
  \includegraphics[width=\columnwidth]{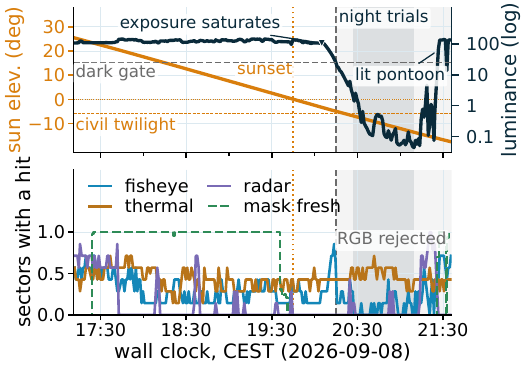}
  \caption{Day-to-night recording from 17:11 to 21:36. Top: solar elevation and RGB luminance. Bottom: per-sensor evidence fractions (60 s medians) and the fraction of onboard updates meeting the mask-age criterion. Gray marks RGB rejection; blue marks night trials.}
  \label{fig:handover}
\end{figure}

Fig.~\ref{fig:handover} shows an offline replay with the deployed scorer, from daylight into night. Automatic exposure maintained image luminance near 105 until reaching its limit at 20:05;
RGB frames fell below the validity threshold approximately ten minutes later. Image luminance masked the decline in ambient light until exposure compensation was exhausted. Once RGB was rejected, thermal and radar continued to provide evidence. Thermal detections persisted across illumination bands, although some arose from water texture, while radar coverage remained limited to its configured \SI{9}{m} region.

\subsection{Ablations}
Table~\ref{tab:ablation} compares sensor evidence, context, and geometric processing on the audited low-light holdout. RGB-only evidence achieves AP~0.903, exceeding the full gradient-boosted model (0.869) when RGB remains informative. Adding fine-tuned detector confidence raises AP to~0.928, with improvements on all eight recordings; the off-the-shelf detector yields a smaller gain (0.890). These results favor informative visual features over additional modalities on this holdout.

Removing context slightly improves AP (0.869 to 0.882) but increases ECE (0.108 to 0.147), indicating a calibration benefit without a ranking benefit. Matching training to sensor availability matters more. A radar-only model achieves AP~0.811 and person recall~1.00, whereas the full model supplied with radar alone achieves AP~0.646 and zero person recall. Explicitly representing missing inputs therefore does not, by itself, ensure reliable predictions under sensor loss.

Removing radar--camera association reduces AP from 0.869 to 0.864 and person recall from 0.89 to 0.86, indicating a small benefit on this set. Removing IMU attitude slightly increases AP to~0.883. Because the segmented water edge supplies the range reference whenever a mask is available, this ablation does not establish whether attitude compensation is useful when that reference is absent. Target-motion features cannot be evaluated on this holdout because no tracked target enters the radar's \SI{9}{m} region.

Separate geometric evaluations support the chosen range reference and sector width. On 310 far-field boat boxes, replacing the IMU-based range reference with the segmented water edge reduces false-near estimates from 12\% to zero. Increasing sector width from $10^{\circ}$ to $15^{\circ}$ improves two-sensor agreement from 78\% to 84\% of frames, at the expense of angular resolution.

\begin{table}[t]
\caption{Scorer ablations on eight audited low-light recordings (2{,}149 sectors, 72\% positive). 
GBT variants share the training split and use OOF calibration unless noted. Typed features use YOLOv8n fine-tuned without holdout data. 
Lower ECE and Brier are better.}
\label{tab:ablation}
\centering
\footnotesize
\setlength{\tabcolsep}{1.8pt}
\begin{tabular}{lccccc}
\toprule
Variant & AP & best F1 & ECE & Brier & Person R \\
\midrule
Rule-based vote & 0.673 & 0.812 & 0.212 & 0.309 & 0.51 \\
Non-negative gate (calibrated) & 0.620 & 0.837 & 0.304 & 0.316 & 0.18 \\
\textbf{GBT (OOF calibration)} & \textbf{0.869} & \textbf{0.837} & \textbf{0.108} & \textbf{0.194} & 0.89 \\
GBT, uncalibrated & 0.879 & 0.803 & 0.227 & 0.265 & 0.89 \\
GBT, train-fit calibration & 0.803 & 0.757 & 0.321 & 0.328 & 0.89 \\
GBT, no context & 0.882 & 0.837 & 0.147 & 0.194 & 0.90 \\
GBT, RGB only & 0.903 & 0.837 & 0.136 & 0.188 & 0.90 \\
GBT, thermal only & 0.703 & 0.836 & 0.209 & 0.280 & 0.61 \\
GBT, radar only & 0.811 & 0.837 & 0.165 & 0.218 & \textbf{1.00} \\
Full GBT, radar-only input & 0.646 & 0.605 & 0.613 & 0.590 & 0.00 \\
GBT, no association & 0.864 & 0.837 & 0.121 & 0.199 & 0.86 \\
GBT, no IMU & 0.883 & 0.837 & 0.126 & 0.192 & 0.90 \\
GBT + off-the-shelf detector (LaRS) & 0.890 & 0.837 & 0.144 & 0.198 & 0.87 \\
\textbf{GBT + on-domain detector} & \textbf{0.928} & \textbf{0.837} & 0.133 & \textbf{0.175} & 0.89 \\
% GBT + target motion & \multicolumn{5}{l}{n/a: no tracked target in holdout} \\
\bottomrule
\end{tabular}
\end{table}

\subsection{Runtime on the Vessel}
\label{sec:runtime}
The Raspberry Pi~4 sustained \SI{3}{Hz} acquisition while running inference separately. On the vessel's supply, the segmentation configuration produced sector estimates at \SI{1.24}{Hz}, with masks within the configured age limit on 99\% of updates. Adding typed detection reduced the rate to \SI{0.57}{Hz}; detector outputs met their age limit on 99\% of updates, but segmentation masks did so on only 5\% (Fig.~S5). When inference fell behind, it processed the newest frame rather than queuing older frames. Consequently, the segmentation configuration skipped 41\% of capture records during the trial evening, while acquisition maintained 900 iterations per five-minute chunk.

\begin{figure}[t]
  \centering
  \includegraphics[width=\columnwidth]{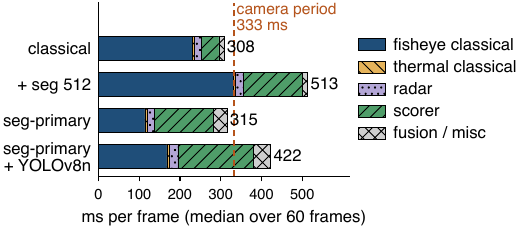}
  % \caption{Per-frame stage budget of the fusion pipeline on the Raspberry Pi~4 at full clock (medians over 60 frames of a busy dockside scene, tick plus scorer; the 333\,ms line is the camera period). The deployed chain adds \SI{0.10}{s} capture-to-disk and \SI{0.13}{s} link latency to the \SI{67}{ms} tail-mode tick: \SI{0.30}{s} from camera frame to the navigation computer.}
  \caption{Raspberry Pi~4 stage timings, including scoring, at full clock (60-frame medians). Dashed line: \SI{333}{ms} camera period.}
  \label{fig:runtimefig}
\end{figure}

Isolated two-thread benchmarks explain these constraints (Fig.~\ref{fig:runtimefig}). Thermal segmentation required \SI{40}{ms} per frame, and the RGB student required \SI{221}{ms} at $512\times384$ resolution, compared with \SI{779}{ms} for its teacher. YOLOv8n/8s inference required 0.68--\SI{4.0}{s}, exceeding the \SI{333}{ms} acquisition period. The classical pipeline with GBT scoring required \SI{308}{ms} at full clock, including \SI{120}{ms} for horizon estimation. Sustained operation was further limited by supply-induced throttling: the processor fell to \SI{600}{MHz}, affecting 48--93\% of frames across runs. The sealed enclosure's idle processor temperature was 73--\SI{78}{\celsius}.

The reduced radar--IMU configuration maintained \SI{3}{Hz} sector updates with \SI{67}{ms} processing time. Its capture-to-navigation latency over Bluetooth LE was \SI{0.30}{s} median and \SI{0.37}{s} at the 95th percentile. Across 24{,}885 transmitted records, no decoding errors or sequence gaps were observed.

% \begin{figure}[t]
%   \centering
%    \includegraphics[width=0.7\columnwidth]{shadow_modes_paper.pdf}
%   \caption{The two on-vessel configurations from their service logs: tick, achieved sector rate and fresh evidence per tick, 5-minute service logs beside capture on the vessel supply (2026-09-07); the board sat at the \SI{600}{MHz} under-voltage clock in both camera configurations.}
%   \label{fig:modes}
% \end{figure}

\subsection{Closed-Loop Pilot}
\label{sec:closedloop}
Each trial began with a \SI{20}{s} observation period with the motor neutral and the rudder and sail centered. The controller computed the median obstacle-evidence score and minimum radar range in each sector, then selected a relative heading within
$\pm45^{\circ}$ by minimizing
\begin{equation}
J(\theta)=\bar q(\theta)
+\lambda_c\,\mathrm{close}(\theta)
+\lambda_h\,|\theta|/\theta_{\max},
\end{equation}
where $\bar q$ is the aggregated score smoothed with a $12^{\circ}$ Gaussian. The closeness term penalizes radar returns within \SI{8}{m}, with a minimum weight of 0.5 regardless of the learned probability; the final term favors forward motion. If $\bar q(\theta^\star)\geq0.5$, the controller continued observation without commanding forward motion. Otherwise, the controller placed a waypoint \SI{15}{m} from the current RTK position along the selected compass heading and used the autopilot's heading controller at fixed low throttle. The initial obstacle-abort rule required a score above 0.5 and a radar return within \SI{4}{m} and $\pm15^{\circ}$ of the course. Loss of communication for \SI{3}{s} also triggered an abort.

Table~\ref{tab:navigation} summarizes 44 armed launches, including 24 driven legs. The first eight legs used the learned probability alone ($q_k=p_k$). Obstacle boats beyond the configured \SI{9}{m} radar region were detected by both cameras ($v_k=0.67$), but the learned probability remained near 0.24. Six of these eight legs selected a straight-ahead heading.

\begin{figure}[t]
  \centering
  \includegraphics[width=\columnwidth]{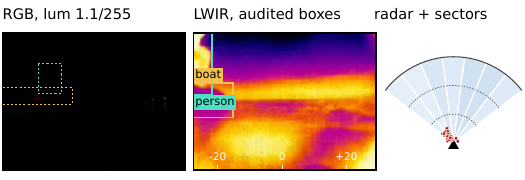}
  \caption{Night example with a crew boat at \SI{1.3}{m}. Left: nearly black RGB image (mean luminance 1.1/255) with audited boxes. Middle: LWIR image with boxes mapped by bearing. Right: radar returns and learned sector probabilities (0.09--0.20), despite a vote of 0.67 where thermal and radar agree.}
  \label{fig:night_triplet}
\end{figure}

From 18:20, the controller used $q_k=\max(p_k,v_k)$. None of the subsequent 16 legs selected straight ahead; the median absolute heading offset was $30^{\circ}$. In 13 legs with a sector at $p_k\geq0.6$, the selected heading was at least $15^{\circ}$ from that sector. The remaining three legs occurred at night, with no sector value above 0.34. RGB frames were rejected as too dark, and heading selection followed small differences in the radar-derived field. Fig.~\ref{fig:night_triplet} presents a separate audited night example in which thermal and radar both detected an obstacle. These sequential trials document how perception affected heading selection.

\begin{table}[t]
\caption{Closed-loop trials with known targets. Ref.: refusals before acceptance; $t_{\mathrm{hdg}}$: median time from launch to heading acceptance; closest: median (minimum) on-course radar range. 
% Evidence: $p_k$ or $\max(p_k,v_k)$. 
% The \SI{4}{m} abort rule triggered once, on the first launch, and was then disabled.
}
\label{tab:navigation}
\centering
\small
\setlength{\tabcolsep}{2.5pt}
\footnotesize
\begin{tabular}{llccccc}
\toprule
Cond. & Evidence & Trials & Ref. & $t_{\mathrm{hdg}}$ (s) & Closest (m) & Aborts \\
\midrule
Day & learned & 12 & 1 & 33 & 10.0 (9.9) & 1 \\
Day & max & 14 & 8 & 34 & 6.7 (3.9) & 0 \\
Dusk & max & 13 & 11 & 20 & 11.5 (3.9) & 0 \\
Night & max & 5 & 0 & 20 & 9.8 (4.0) & 0 \\
\bottomrule
\end{tabular}
\end{table}

\section{Discussion and Conclusion}
We developed and field-tested a sensing-to-navigation pipeline
for a resource-constrained ASV that fuses RGB, LWIR, radar, and inertial
measurements into calibrated sector-level obstacle estimates on a single
onboard computer. It supported closed-loop obstacle avoidance
under outdoor lake conditions with staged targets, including night runs where RGB was
unusable and LWIR and radar supplied the obstacle input. 
Across the held-out low-light recordings, our system produced 
sector probabilities, while LWIR and radar preserved the navigation input
when RGB became unavailable. The non-negative gate exposes modality contributions, but the GBT better models their interactions with range (AP 0.869 vs. 0.620). We hence use the GBT for estimation and the gate for diagnostics. On night recordings with RGB marked unavailable, the rule-based vote performed best. We improved label precision using geometric filtering and a small fine-tuned detector.

%We did not train the full system end-to-end because the available corpus contains only tens of thousands of frames from one lake and 85\% positive sector labels. Adding geographically diverse data raised thermal-segmentation pixel accuracy on the held-out under-pier scene from 0.38 to 0.915, versus 0.97 at the training site. The controller receives explicit sector probabilities, allowing us to trace its actions to the perception output. We log target-motion states separately for evaluating future COLREG-aware planning. The implementation uses a \SI{3}{MB} segmentation model, a sub-megabyte scorer, and one computing board, easing replication across vessels. 

% 1
% Illumination-dependent fusion result
Context-conditioned learned fusion outperformed both the rule-based vote and the interpretable gate on the eight audited low-light recordings, for as long as the RGB images retained useful information. % Table II's twilight column has the vote (0.817) and radar-only (0.822) above the GBT (0.798), so "from twilight into darkness" contradicts our table.
% Main unresolved regime
Night performance is the main unresolved limitation. 
% RGB-unavailable night result
At true night, the fused vote did better on the available audited data, though person recall remains a concern. 
% Probability increments
In the trial evening's shadowed daylight records, fisheye, radar, and thermal votes raised the learned probability. At night, all three added at most 0.04 to a 0.19 baseline, as reproduced offline with the luminance gate. 
% Low thermal weight
The deployed scorer assigned negligible weight to thermal input at all illumination levels, and three of four night navigation decisions used its nearly flat probabilities. 
Site generalization remains limited: geographically diverse training raised thermal-segmentation accuracy on the held-out under-pier scene from 0.38 to 0.915, versus 0.97 at the training site. 
% Retraining result
We retrained the scorer on 247 audited night frames, including 100 with thermal bounding-box annotations, but its night response remained flat. The rule-based fused vote still performed best on these data, with AP 0.58 versus 0.40 for the retrained scorer and 0.52 for the best learned variant; leave-one-recording-out cross-validation preserved this ordering. 

% 3
For audited person detections, the thermal-only and radar-only variants
achieved recalls of 0.90 and 1.00, respectively, versus 0.17 for the
rule-based fused vote. Its higher overall AP therefore masks low person
recall. In one crossing, a wading person at \SIrange{6}{9}{m} received an
obstacle probability of only $p=0.29$ because the blob features did not
distinguish the small target. We added typed detections and motion
estimates, but have not validated them in nighttime crossings within the
radar's \SI{9}{m} range. Learned night fusion requires
substantially more audited frames containing obstacles within that range.

% 4
% Radar operating envelope and camera dependence
We configured the radar for a maximum range of \SI{9}{m}, restricting operation to closing speeds that meet our response-time constraints. 
Beyond this range, estimates rely on cameras, with a baseline obstacle probability of $p\approx0.24$. The thermal camera's narrower field of view leaves the outer sectors radar-only at night, while the sector map omits object shape and may merge objects at similar bearings. We evaluated illumination changes, but not fog, heavy rain, or rough water, and tested closed-loop operation only in the low-speed pilot. The Raspberry Pi~4 limits throughput. A~faster board would increase throughput but would increase compute power draw.

After afloat calibration, monocular range estimates met the $\pm20\%$
acceptance criterion for an isolated target, and radar tracking correctly
classified the evaluated crossings. The on-vessel pipeline produced
sector records at \SI{3}{Hz} with latency below one-third of a second;
the full camera--radar configuration, including typed detection, also ran
alongside capture, albeit at a lower rate. 
% These results support low-speed field operation, while learned fusion under RGB-unavailable night conditions remains the primary validation priority. 
The system therefore enables low-speed field operation, and night-time fusion is the first priority for further validation.

\section*{Acknowledgment}
PK and HH acknowledge DFG support through Germany's Excellence Strategy, EXC 2117--422037984 (CASCB). PK and HH acknowledge the Robotics Institute Germany; ER acknowledges DAAD RISE Germany. PK and ER used large language models solely for language refinement and code-generation assistance, not to generate scientific content.

\section*{Supplementary Material}
Code, evaluation tools, additional figures etc. are available (anonymously): 
\url{https://anonymous.4open.science/r/marine-multimodal-perception-B5EF/}.

\bibliographystyle{IEEEtran}
\bibliography{references_short}

@book{egerstedt2021robot,
  author        = {Egerstedt, Magnus},
  title         = {Robot Ecology: Constraint-Based Design for Long-Duration Autonomy},
  publisher     = {Princeton University Press},
  year          = {2021}
}

@article{bovcon2022wasr,
  author        = {Bovcon, Borja and Kristan, Matej},
  title         = {{WaSR}---A Water Segmentation and Refinement Maritime Obstacle Detection Network},
  journal       = {IEEE Trans. Cybern.},
  volume        = {52},
  number        = {12},
  year          = {2022}
}

@inproceedings{zust2023lars,
  author        = {{\v{Z}}ust, Lojze and Per{\v{s}}, Janez and Kristan, Matej},
  title         = {{LaRS}: A Diverse Panoptic Maritime Obstacle Detection Dataset and Benchmark},
  booktitle     = {Proc. IEEE/CVF Int. Conf. Comput. Vis. (ICCV)},
  year          = {2023}
}

@misc{kedia2024onyx,
  author        = {Kedia, Pranav and Apolinsky, Charlie and Hamann, Heiko},
  title         = {Developing SailSwarm: Small Uncrewed Sailing Vessels for Maritime Environments},
  year          = {2024}
}

@article{tersek2023ewasr,
  author        = {Ter{\v{s}}ek, Matija and {\v{Z}}ust, Lojze and Kristan, Matej},
  title         = {{eWaSR}---An Embedded-Compute-Ready Maritime Obstacle Detection Network},
  journal       = {Sensors},
  volume        = {23},
  year          = {2023}
}

@article{yao2024waterscenes,
  author        = {Yao, Shanliang and others},
  title         = {{WaterScenes}: A Multi-Task 4D Radar-Camera Fusion Dataset and Benchmarks for Autonomous Driving on Water Surfaces},
  journal       = {IEEE Trans. Intell. Transp. Syst.},
  volume        = {25},
  number        = {11},
  year          = {2024}
}

@article{park2024multivessel,
  author        = {Park, Jeong-Ho and others},
  title         = {Multi-Vessel Target Tracking with Camera Fusion for Unmanned Surface Vehicles},
  journal       = {Int. J. Naval Archit. Ocean Eng.},
  year          = {2024}
}

@inproceedings{li2022odombeyondvision,
  author        = {Li, Peize and others},
  title         = {OdomBeyondVision: An Indoor Multi-modal Multi-platform Odometry Dataset Beyond the Visible Spectrum},
  booktitle     = {IEEE/RSJ Int. Conf. Intell. Robots Syst. (IROS)},
  year          = {2022}
}

@inproceedings{jiang2025go,
  author        = {Jiang, Peng and others},
  title         = {GO: The Great Outdoors Multimodal Dataset},
  booktitle     = {IEEE Intell. Veh. Symp. (IV)},
  year          = {2026}
}

@inproceedings{malawade2022hydrafusion,
  author        = {Malawade, Arnav Vaibhav and Mortlock, Trier and Al Faruque, Mohammad Abdullah},
  title         = {{HydraFusion}: Context-Aware Selective Sensor Fusion for Robust and Efficient Autonomous Vehicle Perception},
  booktitle     = {ACM/IEEE 13th Int. Conf. Cyber-Phys. Syst. (ICCPS)},
  year          = {2022}
}

@inproceedings{liu2024groundingdino,
  author        = {Liu, Shilong and others},
  title         = {Grounding DINO: Marrying DINO with Grounded Pre-training for Open-Set Object Detection},
  booktitle     = {Comput. Vis. -- ECCV},
  year          = {2024}
}

@misc{carion2025sam3,
  author        = {Nicolas Carion and others},
  title         = {SAM 3: Segment Anything with Concepts},
  note          = {arXiv:2511.16719},
  year          = {2026}
}

@article{shi2026millimeterwave,
  author        = {Shi, Peixiang and others},
  title         = {Research on an Obstacle Avoidance System for Unmanned Vessels Based on Millimeter-Wave Radar},
  journal       = {J. Mar. Sci. Eng.},
  volume        = {14},
  year          = {2026}
}

@misc{hinton2015distilling,
  author        = {Hinton, Geoffrey and Vinyals, Oriol and Dean, Jeff},
  title         = {Distilling the Knowledge in a Neural Network},
  note          = {arXiv:1503.02531},
  year          = {2015}
}

@inproceedings{howard2019mobilenetv3,
  author        = {Howard, Andrew and others},
  title         = {Searching for {MobileNetV3}},
  booktitle     = {IEEE/CVF Int. Conf. Comput. Vis. (ICCV)},
  year          = {2019}
}

@inproceedings{ke2017lightgbm,
  author        = {Ke, Guolin and others},
  title         = {{LightGBM}: A Highly Efficient Gradient Boosting Decision Tree},
  booktitle     = {Proc. Adv. Neural Inf. Process. Syst. (NeurIPS)},
  year          = {2017}
}

@article{pedregosa2011sklearn,
  author        = {Pedregosa, Fabian and others},
  title         = {Scikit-learn: Machine Learning in {P}ython},
  journal       = {J. Mach. Learn. Res.},
  volume        = {12},
  year          = {2011}
}

@inproceedings{zadrozny2002isotonic,
  author        = {Zadrozny, Bianca and Elkan, Charles},
  title         = {Transforming Classifier Scores into Accurate Multiclass Probability Estimates},
  booktitle     = {Proc. ACM SIGKDD Int. Conf. Knowl. Discov. Data Min.},
  year          = {2002}
}

@inproceedings{guo2017calibration,
  author        = {Guo, Chuan and others},
  title         = {On Calibration of Modern Neural Networks},
  booktitle     = {Proc. 34th Int. Conf. Mach. Learn.},
  volume        = {70},
  year          = {2017}
}
% \bibliography{icra2027_references,paper,article_zotero}
\end{document}